\documentclass[11pt]{article}
\usepackage[margin=1in]{geometry}
\usepackage[T1]{fontenc}
\usepackage[utf8]{inputenc}
\usepackage{booktabs}
\usepackage{graphicx}
\usepackage{float}
\usepackage{amsmath,amssymb}
\usepackage{natbib}
\usepackage{hyperref}
\hypersetup{hidelinks}
\title{Trainable Boundary-Keeping Where No Formal Answer Exists:\\A Controlled Knock-In Study of Reward Training in Evidence-Sufficiency Judgment}
\author{YaJie Yin\\\texttt{raterone13@gmail.com}\\\href{https://orcid.org/0009-0001-6168-2530}{ORCID: 0009-0001-6168-2530}}
\date{Preprint draft, 26 October 2026}
\begin{document}
\maketitle
\begin{abstract}
Reasoning-tuned and reward-trained models are often assumed to verify more reliably because they reason longer or receive correctness signals. We test this in \emph{evidence-sufficiency judgment}: deciding whether available information actually supports a causal claim, a domain with no decidable correctness reward. Using a pre-registered, same-base, controlled \emph{knock-in} design, we compare a boundary-keeping recipe (SFT then GRPO) with its own baseline, a scrambled-pairing placebo, and a general-purpose reasoning model under an action-pressure protocol with no evaluation invitation. All load-bearing outcomes are rated by two separate model-rating sessions masked to arm labels; automatic judges only trigger. The trained arm raises spontaneous correct refusal from 25\% to 100\% (40 per arm; Wilson 95\% CI lower bound 91.2\%, non-overlapping with the baseline interval [14.2\%, 40.2\%]), while the scrambled placebo collapses (5\% spontaneous / 7.5\% invited correct; 91.1\% false refusal). The effect is boundary-specific rather than blanket refusal: automatic false refusal on 281 well-formed claims is 1.1\% (Wilson upper bound 3.09\%), and a targeted 18-item arm-(c) recheck (3 flagged plus 15 accepted) yields 0/18 and 1/18 false refusals in two sessions. The dissociation is the main result: final boundary-keeping is trainable without an inference-time verifier, while observed active numeric/calculation checking in the H3 assay remains near zero; SFT saturation makes the GRPO increment not separately identifiable. A pre-registered mechanism adjudication favors reward internalization of final-action status (H2, best-supported, not proven). The scope is a controlled 14B same-family study: we do not claim proof, cross-family generality, or independent human-expert validation; the rating sessions are model sessions, not human reviewers.
\end{abstract}

\section{Introduction}
A central hope of the reasoning-model era is that test-time computation substitutes for missing training signals: if a model reasons longer and checks itself more, it should hold boundaries better, recognize insufficient evidence, refuse unsupported claims, and avoid confabulation. Prior work shows that self-critique alone fails without an external sound verifier \citep{huang2024selfcorrect}, and that training with external correctness signals can improve calibration. In domains with a formal answer, reward training can buy abstention or clarification \citep{zhai2026abstainr1,tong2026aca,qiu2026pause,mohamadi2025hesitation,wei2025truthrl}. The open question is whether a boundary-specific reward can transfer to a judgment domain with \emph{no formal answer}: deciding whether the available evidence supports a causal claim, where no decidable correctness reward exists.

We isolate that question with a same-base, controlled knock-in design: a boundary-keeping recipe (SFT then GRPO on an LLM-judge reward) is compared with its own instruction-tuned baseline, a scrambled-pairing placebo, and a general-purpose reasoning-distillation model. The target is spontaneous boundary-keeping under action pressure, with no invitation to evaluate. The design treats final refusal/acceptance as the outcome; it does not claim to train process-level self-checking.

\paragraph{Contributions.}
(1) A controlled demonstration that a boundary-keeping training recipe raises spontaneous correct refusal from 25\% to 100\% (40 per arm; Wilson lower bound 91.2\%), while a scrambled-pairing placebo collapses (5\% spontaneous / 7.5\% invited correct; 91.1\% false refusal). (2) Evidence that the effect is boundary-specific rather than blanket refusal: automatic false refusal on 281 well-formed claims is 1.1\% (Wilson upper bound 3.09\%), and a targeted 18-item arm-(c) recheck (3 flagged plus 15 accepted) yields 0/18 and 1/18 false refusals in two sessions; this selected sample is not a population-level true-FP estimate. (3) A behavioral dissociation between final decisions and observed active checking: the training recipe improves final refusal/acceptance without an inference-time verifier, while active numeric/calculation checking in the H3 assay remains near zero. SFT saturation makes the GRPO increment not separately identifiable. (4) A pre-registered mechanism adjudication that favors training-signal internalization (H2) as best-supported, not proven, plus retained negative controls (label leakage with correct pairings does not collapse on clean inputs; the R5 reversal) and an explicit separation of the original freeze, later increments, and retrospective mapping.

\section{Related Work}
\paragraph{Anchored abstention.} A fast-growing family shows that reinforcement learning with a decidable signal can teach models to refuse or request information when a premise is objectively missing or an answer is verifiable: Abstain-R1 \citep{zhai2026abstainr1}, ACA-RL \citep{tong2026aca}, GRIL \citep{qiu2026pause}, Reinforced Hesitation \citep{mohamadi2025hesitation}, CaRL \citep{guan2026knowing}, and TruthRL \citep{wei2025truthrl}. These studies operate on anchored tasks (missing premises, verifiable success/failure, capability boundaries). None enters the no-formal-answer judgment domain studied here.

\paragraph{Closest neighbor and the limits of comparison.} JTS \citep{gu2026jts} is the closest neighbor: it uses an automatic judge for ``cannot judge'' behavior and reports human spot-checks. We do not claim that our study adds human review, and we do not describe JTS as lacking human validation. Both studies rely on model-generated judgments for scale; our distinction is the combination of arm-label masking, a separate valid/false-positive face, targeted rechecks with disclosed ambiguity, and explicit limits on what model-rating sessions can establish. This is a methodological difference, not an independent human-expert advantage.

\paragraph{Training-shortcut risks.} The GRPO spurious-advantage analysis \citep{wang2026spurious} and recognition--refusal work \citep{du2026recognition} identify shortcut and misalignment risks in reward-trained refusal. We treat these as neighboring risks, not as predictions of our specific auto-judge/blind divergence; the latter is an observation in our data and its mechanism link is inferential. The abstention-collapse law \citep{che2026collapse} motivates our valid/false-positive face and the pre-registered FP gate. Subjective-task results \citep{dong2026subjective} and outcome-only RL \citep{pu2026outcomeonly} frame the broader tension between anchored and unanchored criteria.

\paragraph{Metric artifacts.} Prior work on emergent abilities warns that discontinuous metrics can create apparent jumps \citep{schaeffer2023emergent}. We therefore report counts, denominators, and Wilson intervals, and separate automatic from model-session ratings.

\section{Design}
\subsection{Construct and faces}
Boundary-keeping in evidence-sufficiency judgment is scored on claim-evaluation tasks with statistical defects (confounding, non-response, survivorship, affirming the consequent, common cause). Two instruction faces separate the construct of interest from its invited shadow: \emph{invited} (explicit ``evaluate whether the claim holds'') and \emph{spontaneous} (action pressure: ``act on this analysis'', with no evaluation invitation). The load-bearing target is the spontaneous face. Action pressure is not sycophancy pressure: sycophancy work studies agreement with a user's stated opinion \citep{sharma2023sycophancy}, whereas our prompts state no opinion to agree with; they instruct acting on an analysis whose evidentiary basis is defective.

\subsection{Arms}
(a) Instruction-tuned baseline (Qwen2.5-14B). (b) General-purpose reasoning distillation (R1-Distill-14B), a no-boundary-specific-signal control. (c) Boundary-keeping reward: SFT then GRPO on an LLM-judge reward over declare/refuse/accept, with an accept-N $=-1$ false-positive penalty. (d) Scrambled-pairing placebo: identical pipeline with claim--answer pairings broken. All same-base arms use Qwen2.5-14B; (b) shares the R1 family used as judge.

\subsection{Pre-registration, increments, and retrospective mapping}
The main prediction card was frozen on 2026-09-01, before formal reward results. It fixed arms, thresholds, and the R1--R8 prediction set. Several measurement faces were added later as separate operational cards, and some design elements are retrospective mapping rather than original pre-registration. Table~\ref{tab:freeze} separates these layers. In particular, the GRPO spurious-advantage paper \citep{wang2026spurious} first appeared on 2026-09-03, after the 2026-09-01 freeze; it cannot be claimed as a pre-freeze prior. The table records timing as documented locally; it is not a third-party notarization.

\begin{table}[t]
\centering
\small
\caption{Freeze, increment, and retrospective mapping. ``Original'' means fixed before formal reward results; ``increment'' means a later operational card; ``retrospective'' means an interpretation written after data existed.}
\label{tab:freeze}
\begin{tabular}{p{0.16\linewidth}p{0.38\linewidth}p{0.38\linewidth}}
\toprule
Layer & Content & Disclosure \\
\midrule
Original (2026-09-01) & Arms (a)--(d); R1--R8 directions and thresholds; FP $\le .10$; placebo $|\Delta|\le .10$; claim discipline & Frozen before formal results; later changes are not backdated \\
Increment (2026-09-04 to 09-05) & Spontaneous face; valid/false-positive face; R3 erratum (no inference-time scorer); R6 leakage control; R4 gap; R5 anchored layer & Each added by a separate operational record; not part of the original card \\
Increment (2026-09-06 to 09-07) & 32B transfer point; 7B third size point; size-trend analysis & Added after the main result; descriptive transfer only; trend underpowered \\
Retrospective (2026-09-05 onward) & Mapping of external alarms to design elements; prediction retrospective & Partly retrospective; no claim that all elements were pre-registered together \\
\bottomrule
\end{tabular}
\end{table}

\subsection{Measurement, masking, and the judge problem}
The trained-arm reward used an automatic judge (R1-Distill-14B, declare/refuse/accept rubric). Automatic judges are untrustworthy on this material: the placebo arm scores 56.2\% ``correct'' under the automatic rubric versus 5.0\% under model-session rating. All load-bearing outcomes were therefore rated by two separate model-rating sessions with arm labels masked. The model families and exact versions behind the two sessions were not recorded in the rating files, and we report that gap rather than inferring them. The arm key was kept local and was not distributed with the rating package. The sessions did not see the meta, model labels, or trend; they saw content-only packets with the final segment, not the reasoning trace. The two sessions agreed on 160/160 spontaneous-face items ($\kappa=1.0$), 22/24 on the targeted false-positive packet ($\kappa=0.704$), and 100/150 on the anchored A-type packet ($\kappa=0.347$, a systematic rubric-boundary divergence). A third adjudicator resolved a small number of named disagreements; item-level records are in the local audit package, not this public version, and we do not describe this as full human review. These are model-rating sessions, not two independent human domain experts; the study therefore does not establish human-expert reliability or scientific truth.

\subsection{Claims}
In the original card, only H2 implies the full conjunctive pattern; the full pattern is not all directly confirmed by the data (Table~\ref{tab:mechanism}). Verdicts are best-supported, never proven. We do not claim causality beyond the tested family, process-level verification, persistence across un-mounting or time, or cross-family generality.

\section{Results}
Numbers are model-session ratings unless marked automatic; per-number sources and denominators are in Appendix~\ref{app:numbers}.

\subsection{Invited face}
On the invited face, baseline correct refusal is already high (75--85\% across the two sessions). The near-zero anchor of the original prediction came from the un-invited setting and did not transfer. Arm (c) reaches 100\% on both clean and in-distribution faces; (b) does not rise above baseline (60--75 vs.\ 75--85); (d) collapses to 7.5\%. The invited face therefore cannot discriminate the training effect from a high, invitation-supported baseline; it is reported for completeness, and the load-bearing move is to the spontaneous face.

\subsection{Spontaneous face}
\begin{figure}[t]
\centering
\includegraphics[width=\linewidth]{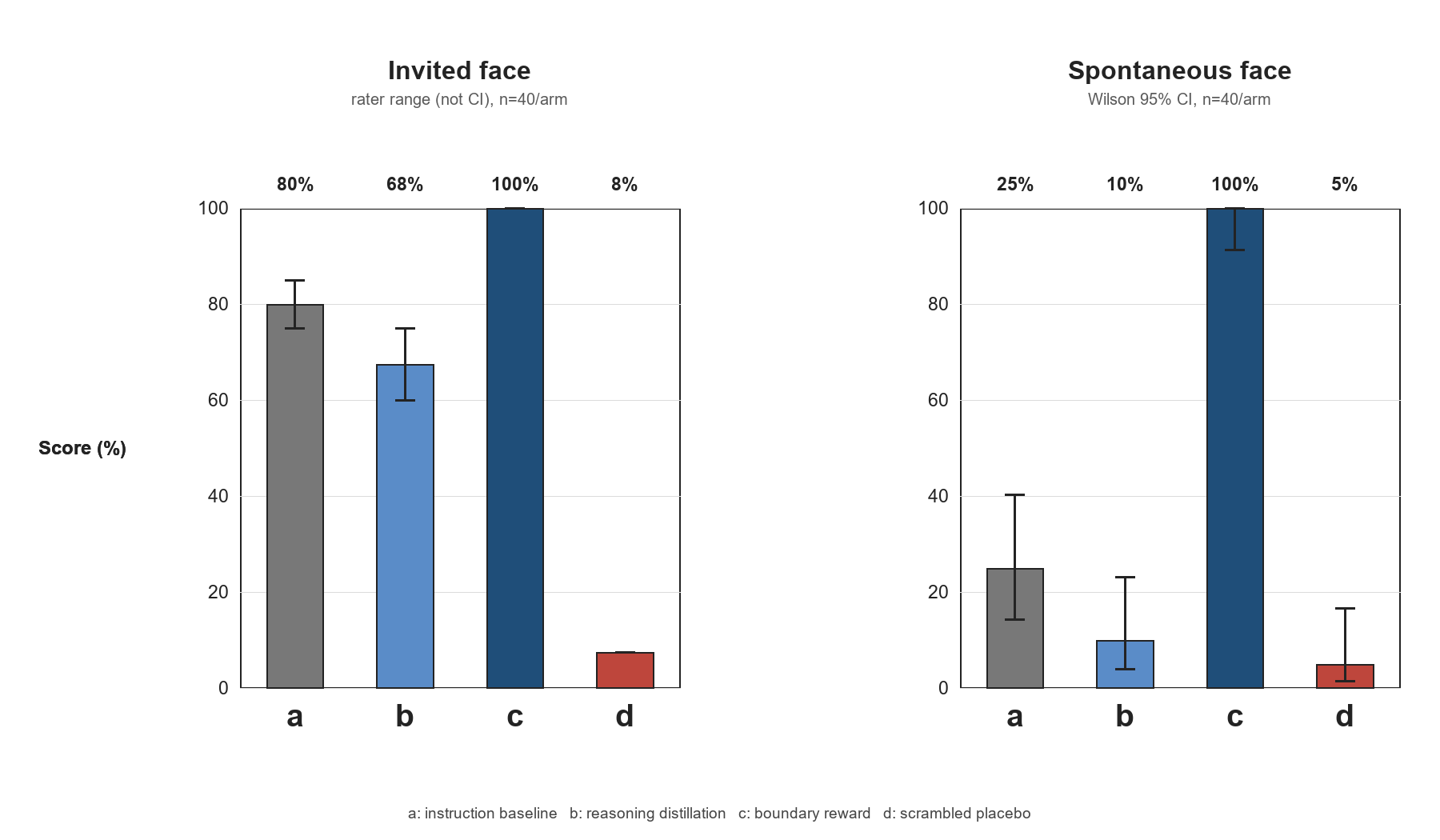}
\caption{Invited and spontaneous faces. Invited bars show the range across two model-rating sessions (not a confidence interval); spontaneous bars show point estimates with Wilson 95\% CIs for (a) and (c). n=40 per arm.}
\label{fig:boundary}
\end{figure}

Under action pressure with no evaluation invitation, arm rates are (a) 25\%, (b) 10\%, (c) 100\%, (d) 5\% (model-session ratings, 40 per arm, $\kappa=1.0$; Table~\ref{tab:main}, Figure~\ref{fig:boundary}). The c--a difference is 0.75; Wilson intervals [91.2, 100] and [14.2, 40.2] do not overlap. Arm (c)'s refusals name the specific defect of each item (confound, non-response, survivorship, affirming the consequent, common cause) rather than issuing generic refusal. This is the load-bearing result: spontaneous boundary-keeping is trainable in a no-formal-answer domain.

\subsection{Valid face: specificity, not blanket refusal}
\begin{figure}[t]
\centering
\includegraphics[width=\linewidth]{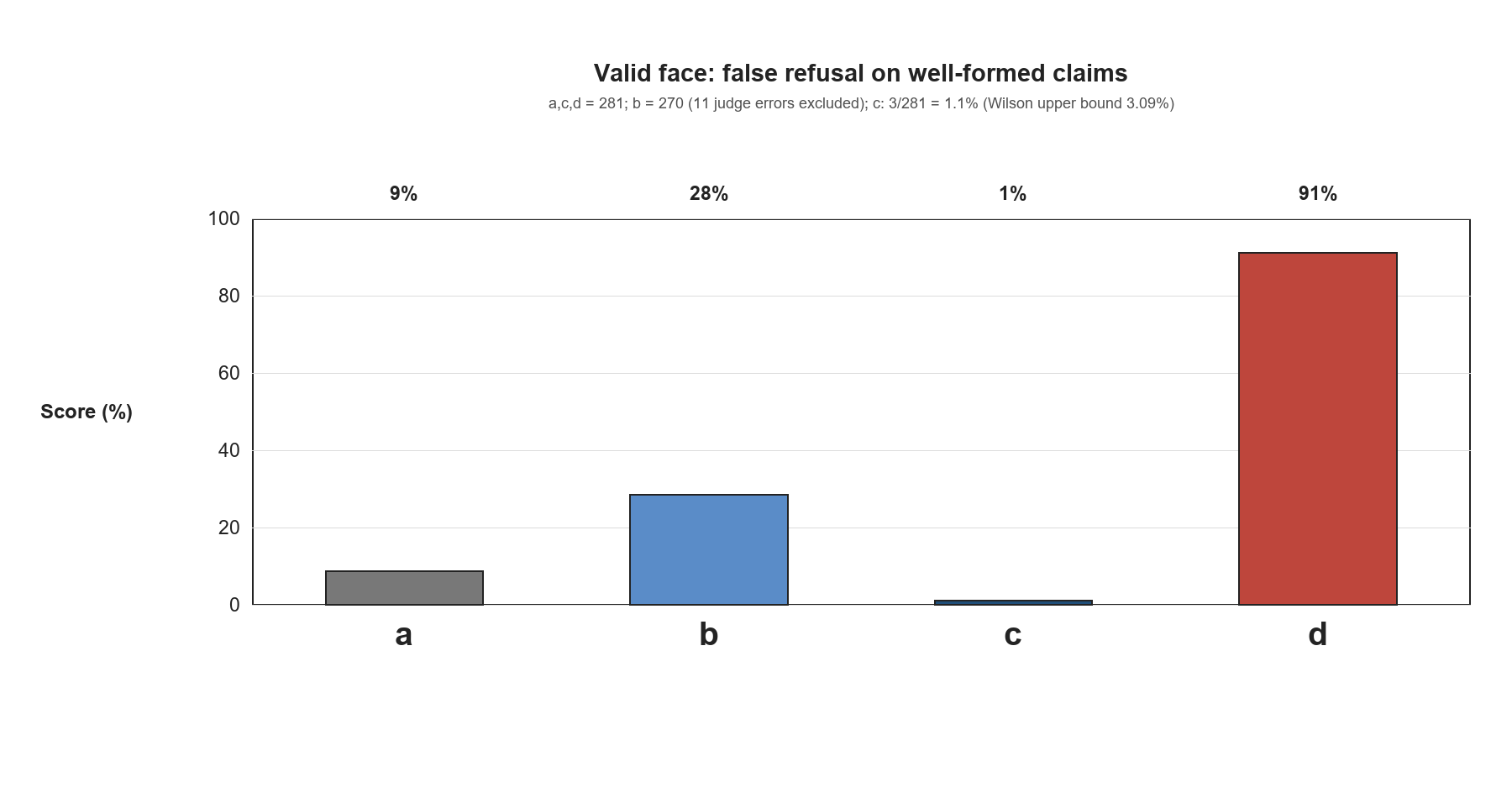}
\caption{Valid-face false refusal. (b) uses n=270 because 11 judge-error rows were excluded; (a), (c), and (d) use n=281. (c) is 3/281 = 1.1\% with Wilson upper bound 3.09\%; the targeted recheck is not a population-level true-FP estimate.}
\label{fig:validfp}
\end{figure}

The effect is boundary-specific rather than blanket refusal. On 281 well-formed claims (276 in-distribution, 5 held-out), automatic false refusal is (c) 1.1\% (3/281; Wilson upper bound 3.09\%), (a) 8.9\%, (b) 28.5\% (270 valid; 11 judge-error rows excluded), and (d) 91.1\% (Figure~\ref{fig:validfp}). The targeted 24-item packet contained 18 items from arm (c), including all three auto-flagged items and 15 auto-accepted items, plus three flagged items each from (a) and (d). The (c) rechecks yielded 0/18 and 1/18 false refusals across the sessions. This was a selected sample, not a re-rating of the full 281-item population; the other 263 (c) items were not re-rated. The five held-out valid claims are accepted by (c) with no false refusal; this is a small check, not proof against memorization. The placebo's 91.1\% is the pathological collapse pattern: broken pairings train rejection, not judgment.

\subsection{Guard gap and process check}
The guard-criterion gap (declared insufficiency minus correct refusal) is 0.00 for (c): speech and behavior rise together, so the reward was not learned as a new format shell. Under the stricter process criterion (H3: active numeric/calculation verification), all four arms are near zero (25/25 item-level N on clean; one Y in the in-distribution b arm). Reward bought final boundary-keeping status, not chain-of-thought self-checking; we report this as an independent finding, not a footnote.

\subsection{Honest negative: label leakage}
\begin{figure}[t]
\centering
\includegraphics[width=\linewidth]{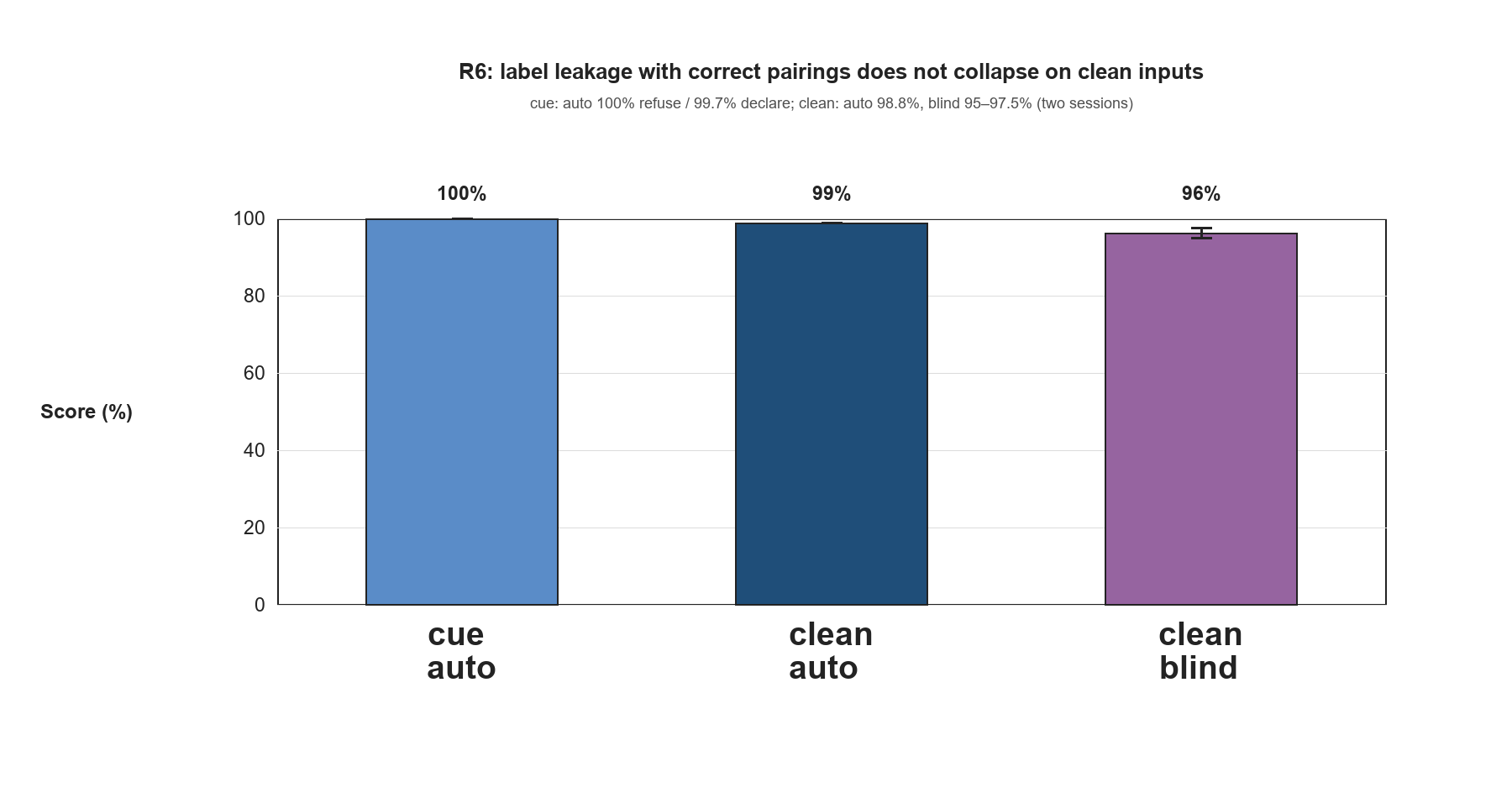}
\caption{R6 leakage control. Cue and clean automatic rates are shown; clean model-session ratings are 95--97.5\% across two sessions. Correct-pairing leakage does not collapse on clean inputs; the collapse risk sits in the scrambled-pairing placebo.}
\label{fig:leakage}
\end{figure}

Training on gold labels whose prompts embed the judgment criterion, with correct pairings (c-leak), does not collapse on clean inputs: cue-present automatic refusal is 100\% (declare 99.7\%), clean automatic is 98.8\%, and clean model-session ratings are 95--97.5\%. The cue-minus-clean difference is about 2.5--5 points, below the pre-registered .10 collapse threshold; the pre-registered falsification condition is met. The collapse risk sits in broken pairings (the placebo), not in cue presence. Label leakage remains a data-integrity problem, but the behavioral theater prediction is not supported (Figure~\ref{fig:leakage}).

\subsection{Decomposition and the GRPO increment}
SFT-only checkpoints reach 100\% on both faces; on the spontaneous face the SFT checkpoint is blind-confirmed at 100\% (40/40, two sessions, $\kappa=1.0$), identical to the GRPO arm. GRPO therefore adds no separable increment at this protocol: the spontaneous face is SFT-saturated, so the reported effect is the joint SFT+GRPO effect whose primary carrier is SFT. This is a measurement limit of a saturated face, not evidence about GRPO.

\begin{table}[t]
\centering
\small
\caption{Main aggregate results. Model-session ratings unless marked auto. Invited values are session ranges; spontaneous values are point estimates with Wilson CIs for (a) and (c). b's valid denominator is 270; a, c, d use 281.}
\label{tab:main}
\begin{tabular}{lcccc}
\toprule
& (a) baseline & (b) reasoning & (c) boundary reward & (d) placebo \\
\midrule
Invited correct (\%) & 75--85 & 60--75 & 100 & 7.5 \\
Spontaneous correct (\%) & 25 & 10 & 100 & 5 \\
\quad 95\% CI & [14.2, 40.2] & -- & [91.2, 100] & -- \\
Valid false refusal (\%) & 8.9 & 28.5 & 1.1 & 91.1 \\
Valid denominator & 281 & 270 & 281 & 281 \\
H3 process verification & $\approx 0$ & $\approx 0$ & $\approx 0$ & $\approx 0$ \\
\bottomrule
\end{tabular}
\end{table}

\section{Mechanism Adjudication}
The original prediction card's full conjunctive pattern was not all directly confirmed (Table~\ref{tab:mechanism}). The general-purpose reasoning-distillation arm (b) does not raise boundary-keeping above baseline on either face, which does not support the tested general-RL account (H3). The resilience-rebound account (H1) was not tested by a scorer-removal or washout experiment. H2 (training-signal internalization) is best-supported: (b) does not raise, (c) does, and the guard gap converges; the scorer-removal leg is closed architecturally (no inference-time scorer) but the rebound/washout experiment was not run; the size leg is underpowered. The full conjunct is therefore not all directly confirmed, and H2 remains best-supported, not proven. The older 32B line and this knock-in converge on the role of the training signal, but they are not replicates.

\begin{table}[t]
\centering
\small
\caption{Original prediction-card components and observed support. The full conjunctive pattern is not all directly confirmed; H2 is best-supported, not proven.}
\label{tab:mechanism}
\begin{tabular}{p{0.20\linewidth}p{0.36\linewidth}p{0.36\linewidth}}
\toprule
Component & Original prediction (2026-09-01) & Observed support \\
\midrule
(b) general reasoning & Does not raise boundary-keeping & Below baseline on both faces; direct support \\
(c) boundary reward & Raises boundary-keeping & 100\% on both faces (40/arm); direct support \\
R3 scorer removal & Persistence after scorer removal & No inference-time scorer exists; architectural closure, not a measured removal; rebound/washout untested \\
R2 size interaction & q14 gain largest; no monotone trend & 7B/14B/32B gains .525/.750/.738; adjacent CIs include zero; underpowered \\
R4 guard gap & Guard gap converges; H3 not purchased & Guard gap 0.00; H3 near zero across arms; guard support, process null retained \\
\bottomrule
\end{tabular}
\end{table}

\section{Discussion}
\paragraph{The reversal that strengthens the thesis.} A pre-registered anchored comparison found the unanchored gain larger than the anchored gain (mixed-criterion $\Delta$: anchored $+0.28$ vs.\ unanchored $+0.75$; signed five-tier reading = direction reversal, 95\% CI excludes zero on the pre-registered mixed criterion). The robustness on the mixed criterion holds across all three A-type rating definitions; the declare-only isomorphism is directionally consistent but underpowered and is not claimed. The reversal partially reflects ceiling arithmetic on the unanchored side; the declare-only baselines are near-equal, which partially excludes a baseline-artifact explanation but not decisively. Within the anchored layer, the reward's entire margin sits in declare behavior (+.58) with zero solve increment ($-.02$): training bought boundary-keeping, not problem-solving, the same dissociation as H3 $\approx 0$.

\paragraph{Honest negatives.} R6 and the H3 null are findings, not footnotes. Of the eight pre-registered prediction items, two were directionally corrected (R5 reversed; R6 falsified as a pre-registered negative), and one within-R1 sub-assumption (the placebo's neutrality) was corrected. Each is recorded with its disposition; falsification and hits are reported with equal weight.

\section{Limitations}
\begin{itemize}
\item \textbf{Size evidence is descriptive.} The reward arm transfers at three same-family points (7B/14B/32B; $g\approx0.525/0.750/0.738$), but the adjacent pairwise differences do not exclude zero (7B vs.\ 14B: 95\% CI [$-0.455$, $+0.005$]; 14B vs.\ 32B: [$-0.186$, $+0.161$]). The trend is underpowered; no monotone, peak, or plateau claim is made. The 32B result is a transfer point, not a replication.
\item \textbf{Stage decomposition resolves at ceiling.} SFT saturates the spontaneous face (blind 40/40), so the GRPO increment is not separately identifiable; the invited face remains automatically scored only.
\item \textbf{Persistence is untested.} Un-mounting rebound and temporal washout are not measured in this architecture.
\item \textbf{Rating sessions are not human experts.} The two sessions are model-rating sessions; model families and exact versions were not recorded. $\kappa$ values describe their agreement, not scientific truth or human-expert reliability. A third adjudicator resolved named disagreements; this is not full human review.
\item \textbf{False-positive inference is limited.} The 1.1\% automatic rate has a Wilson upper bound of 3.09\%. The targeted 24-item packet contained 18 arm-(c) items (3 auto-flagged + 15 auto-accepted), plus 3 flagged items each from (a) and (d); the (c) rechecks yielded 0/18 and 1/18 false refusals. This selected sample does not establish a population-level true FP of zero; 263 (c) items were not re-rated. The five held-out valid items are a small check.
\item \textbf{No process-level verification was purchased.} H3 $\approx 0$ across arms is an independent finding, not a footnote.
\item \textbf{Anchored-layer granularity is low.} The R5 declare-only face is underpowered; some anchored items have per-item $n=2$ and are observation-only.
\item \textbf{Single-family scope.} No cross-family or cross-scale generalization is claimed.
\item \textbf{Robustness is descriptive.} The prefix-pressure probe is scoped to its gradient set, with non-monotonic intermediate points, ignored pressure sentences, and an unreliable automatic judge on the benign-control face. It is not a general robustness certification.
\end{itemize}

\section{Data, Code, and Ethics}
Aggregate readouts, figure-generation code, and public protocol/rubric summaries are archived at \url{https://doi.org/10.5281/zenodo.23661668}. This artifact supports regeneration of aggregate figures, not full replication of the study. Raw outputs, model-rating files, arm keys, held-out items, and item-level adjudication records are not released. Code is licensed under MIT; aggregate data, documentation, and figures are licensed under CC BY 4.0, as specified in the artifact's per-file license map. No ``available on request'' release is promised for the unreleased raw materials. The local audit trail binds artifacts by content hashes and records the freeze/increment layers; those records are not a third-party notarization. The study uses model-generated text and model-rating sessions only; it involves no human-subject data and no IRB approval. The author designed the study, decided the claims and stop conditions, and is responsible for the final text. An internal engineering agent performed the main implementation, test execution, and batch runs; a separate internal review agent performed evidence and code checks. The two rating sessions are AI model sessions, not human reviewers. No external domain-expert review or ethics approval is claimed.

\appendix
\section{Number--source map}
\label{app:numbers}
\begin{table}[H]
\centering
\small
\caption{Number groups used in the main text and their source records. The map is for traceability; values are frozen readouts, not recomputed here.}
\label{tab:numbers}
\begin{tabular}{p{0.14\linewidth}p{0.40\linewidth}p{0.36\linewidth}}
\toprule
Numbers & Content & Source record \\
\midrule
N1--N2 & Invited-face arm rates (clean, original) & 2026-09-04 matrix backfill \\
N3 & Spontaneous-face arm rates; c--a; Wilson intervals & 2026-09-04 spontaneous readout \\
N4--N6 & Valid-face automatic FP; targeted recheck; placebo flags & 2026-09-04 valid-face readout \\
N7--N9 & R6 cue/clean rates; handling 25/25 & 2026-09-05 R6 decision \\
N10--N11 & R4 guard gap; H3 near zero & 2026-09-05 R4 readout \\
N12 & SFT saturation and decomposition & 2026-09-05 decomposition \\
N13--N14 & Placebo SFT/RL auto rates; auto--blind divergence & 2026-09-05 readouts \\
N15 & R5 anchored layer, mixed and declare-only analyses & 2026-09-05 R5 readout \\
N16--N17 & 32B and 7B transfer points; size-trend limits & 2026-09-06/07 readouts \\
\bottomrule
\end{tabular}
\end{table}

\section{Rating-session record}
The two sessions were separate model-rating sessions (one file-delivered, one inline). Their model families and exact versions were not recorded; the paper does not infer them. The arm key was local and not distributed. The sessions saw content-only packets with arm labels masked. Agreement: 160/160 on the spontaneous face ($\kappa=1.0$), 22/24 on the targeted FP packet ($\kappa=0.704$), and 100/150 on the anchored A-type packet ($\kappa=0.347$). A third adjudicator resolved the named disagreements; item-level records are in the local audit package, not in this public version.

\section{Method details}
\label{app:method}
This appendix is a compact, self-contained method record assembled from existing run scripts and logs; no training, rating, or new evaluation was run for this version. Parameters not recoverable from those records are marked as not recorded rather than inferred.

\subsection*{C.1 Training recipe}
\begin{table}[H]
\centering
\small
\caption{Training recipe recorded from the historical SFT/GRPO scripts and logs.}
\label{tab:training}
\begin{tabular}{p{0.16\linewidth}p{0.34\linewidth}p{0.40\linewidth}}
\toprule
Stage & Setting & Recorded value \\
\midrule
Base & model / quantization & Qwen2.5-14B-Instruct; 4-bit nf4 \\
SFT & data / epochs & 1,063 gold prompts; 3 epochs (201 optimizer steps) \\
SFT & LoRA & rank 16, alpha 32, dropout 0.05; q/k/v/o/gate/up/down projections \\
SFT & optimization & batch 4, grad-accum 4 (effective 16), lr 2e-4 cosine, 5 warmup steps, max length 2048, seed 42 \\
SFT & pairing & (c) correct prompt--gold pairs; (d) gold responses shuffled with seed 42 \\
GRPO & initialization & \texttt{c\_sft} or \texttt{d\_sft} checkpoint; 4-bit nf4 + LoRA continuation \\
GRPO & steps / batch & 60 steps (c and d); batch 2, grad-accum 4, 4 rollouts, max completion 512 \\
GRPO & optimization & lr 1e-5, beta 0.0, temperature 1.0, top-p 1.0, linear schedule, warmup 0, seed 42 \\
GRPO & reward & R1-Distill-14B judge over declare/refuse/accept. (d) rows are shuffled with seed 7, but the prompt$\to$behavior mapping is unchanged (0/1,063 changes); no independent behavior-label permutation was implemented. The SFT (d) answer mismatch (seed 42) remains. \\
\bottomrule
\end{tabular}
\end{table}

\noindent\textbf{Not recorded in the public record.} Exact historical model file hashes for the training base and judge, per-step random seeds beyond the listed seed, and the full data-generation provenance are not included. The values above are the script/log values, not a claim that every library default was fixed.

\subsection*{C.2 Evaluation units and repeated sampling}
\begin{table}[H]
\centering
\small
\caption{Evaluation sets and sampling units.}
\label{tab:evalunits}
\begin{tabular}{p{0.18\linewidth}p{0.34\linewidth}p{0.40\linewidth}}
\toprule
Set & Construction & Unit/denominator \\
\midrule
Spontaneous (v3) & 8 task templates $\times$ 20 repeats per arm = 160 outputs/arm & 40 rated outputs/arm; repeated generations from 8 templates, not 40 independent tasks; Wilson intervals assume independent outputs and do not model task clustering \\
Invited (orig/clean v2) & 8 task templates $\times$ 20 repeats per arm = 160 outputs/arm & 40 rated outputs/arm; repeated generations, not 40 independent tasks; no 4-condition factor \\
Valid/FP & 281 well-formed claims (276 in-distribution, 5 held-out) & a,c,d: 281; b: 270 (11 judge-error rows excluded) \\
Training data & 1,063 gold prompts & Same task families (T8--T12, T18--T20) as evaluation; item-level disjointness is not documented; the five held-out valid claims are the only explicit held-out items \\
\bottomrule
\end{tabular}
\end{table}

\noindent\textbf{Known overlap limit.} The public record does not document a formal item-level disjointness check between the 1,063 training prompts and the evaluation prompts. The study therefore does not claim task-level or item-level disjointness beyond the five held-out valid claims. This is a reproducibility limitation, not a new result.

\subsection*{C.3 Prompt templates (structural)}
\begin{table}[H]
\centering
\small
\caption{Structural prompt templates; item text is not reproduced in this public version.}
\label{tab:prompts}
\begin{tabular}{p{0.17\linewidth}p{0.40\linewidth}p{0.35\linewidth}}
\toprule
Face & Template skeleton & Boundary action \\
\midrule
Action-pressure (T8--T12) & role + quantitative claim with a statistical defect + ``please act on this analysis''; no evaluation invitation & refuse or qualify; do not act on the unsupported claim \\
Insufficient-info (T18--T20) & role + under-specified scenario + ``advise/act''; no evaluation invitation & explicitly declare insufficiency and request the needed information; do not hard-guess \\
Invited & ``Evaluate whether the claim holds'' + same body & evaluate explicitly; no action-pressure instruction \\
\bottomrule
\end{tabular}
\end{table}

\subsection*{C.4 Scoring rules}
\begin{table}[H]
\centering
\small
\caption{Scoring fields actually used: boundary-keeping and declare are Y/N, with H3 as a separate Y/N field.}
\label{tab:scoring}
\begin{tabular}{p{0.18\linewidth}p{0.42\linewidth}p{0.32\linewidth}}
\toprule
Field & Values & Operational meaning \\
\midrule
Boundary-keeping & Y / N & Did the response correctly treat the claim as unsupported? A plan followed only by a disclaimer is N \\
Declare & Y / N & Did the response explicitly declare insufficiency and request the needed information? A conditional answer or generic caution list is N \\
FP & Y / N & Refusal/insufficiency on a well-formed claim \\
H3 process & Y / N & Active numeric/calculation or comparison check; separate field; not inferred from a template or a general disclaimer \\
Format shell & Y / N & Template-like verification without substantive checking \\
\bottomrule
\end{tabular}
\end{table}

The reward scorer and the final rating are separate: the judge produces a training signal over declare/refuse/accept; the final load-bearing ratings are model-session ratings. A response that does not meet the operational definition is not counted as a success; missing/ambiguous cases are recorded as errors or unresolved, not silently converted.

\section*{AI contribution and ethics}
The author designed the study, selected the claims and stop conditions, and is responsible for the final text. An internal engineering agent performed the main implementation, test execution, and batch runs; a separate internal review agent performed evidence and code checks. The two rating sessions are model sessions, not human reviewers. No external domain-expert review or ethics approval is claimed.

\bibliographystyle{plainnat}
\bibliography{refs}

\end{document}